\documentclass[letterpaper, 10 pt, conference]{ieeeconf}  

\IEEEoverridecommandlockouts                              

\usepackage{graphicx} 
\usepackage{epsfig} 
\usepackage{mathptmx} 
\usepackage{times} 
\usepackage{amsmath} 
\usepackage{amssymb}  
\usepackage{bm}
\usepackage{subfig}
\usepackage{url}
\usepackage{color}
\DeclareMathOperator*{\argmin}{argmin}
\title{\LARGE \bf
Dynamic Reconstruction of Deformable Soft-tissue
with Stereo Scope in Minimal Invasive Surgery
}

\author{Jingwei Song$^{}$, Jun Wang$^{}$, Liang Zhao$^{}$, Shoudong Huang$^{}$ and Gamini Dissanayake$^{}$
\thanks{$^{}$All the authors are from Centre for Autonomous Systems, University of Technology, Sydney, P.O. Box 123, Broadway, NSW 2007, Australia}%
\thanks{Email: jingwei.song@student.uts.edu.au, wangjun@radi.ac.cn, \{Liang.Zhao; Shoudong.Huang;
Gamini.Dissanayake\}@uts.edu.au} 
}

\begin{document}

\maketitle
\thispagestyle{empty}
\pagestyle{empty}

\begin{abstract}

In minimal invasive surgery, it is important to rebuild and visualize the latest deformed shape of soft-tissue surfaces to mitigate tissue damages. This paper proposes an innovative Simultaneous Localization and Mapping (SLAM) algorithm for deformable dense reconstruction of surfaces using a sequence of images from a stereoscope.  We introduce a warping field based on the Embedded Deformation (ED) nodes with 3D shapes recovered from consecutive pairs of stereo images. The warping field is estimated by deforming the last updated model to the current live model. Our SLAM system can: (1) Incrementally build a live model by progressively fusing new observations with vivid accurate texture. (2) Estimate the deformed shape of unobserved region with the principle As-Rigid-As-Possible. (3) Show the consecutive shape of models. (4) Estimate the current relative pose between the soft-tissue and the scope. In-vivo experiments with publicly available datasets demonstrate that the 3D models can be incrementally built for different soft-tissues with different deformations from sequences of stereo images obtained by laparoscopes. Results show the potential clinical application of our SLAM system for providing surgeon useful shape and texture information in minimal invasive surgery.

\end{abstract}

\section{INTRODUCTION}

Minimally Invasive Surgery (MIS) is an indispensable tool in modern surgery for the ability of mitigating postoperative infections, but it also narrows the surgical field of view and makes surgeons receive less information. MIS has introduced significant challenges to surgeons
as they are required to perform the procedures in narrow space with elongated tools without direct 3D vision \cite{mountney2009dynamic}. To solve this problem, 3D laparoscopy is applied to provide two images to create an `imagined 3D model' for surgeons. Inspired by the fact that stereo vision can generate shapes for qualitative and quantitative purpose, mosaic all the 3D shape by taking account deformation will make better use of 3D information. Therefore, it is helpful if a dynamic 3D morphology could be incrementally generated and rendered for the surgeons intra-operatively and for future autonomous surgical robots (known as computer assisted intervention) for implementing surgical operation and navigation. However, the small field of view of the scopes and the deformation of the soft-tissue limit the feasibility of using traditional structure-from-motion and image mosaicking methods. Even worse, rigid and non-rigid movement caused by motion of camera pose, breathing, heartbeat and instrument interaction increase difficulty in soft-tissue reconstruction and visualization. Therefore, in this paper, we focus on incrementally recovering the morphology and motion of soft-tissues with stereoscope intra-operatively. \par
\begin{figure}[]
	\centering
	\includegraphics[width=0.5\textwidth]{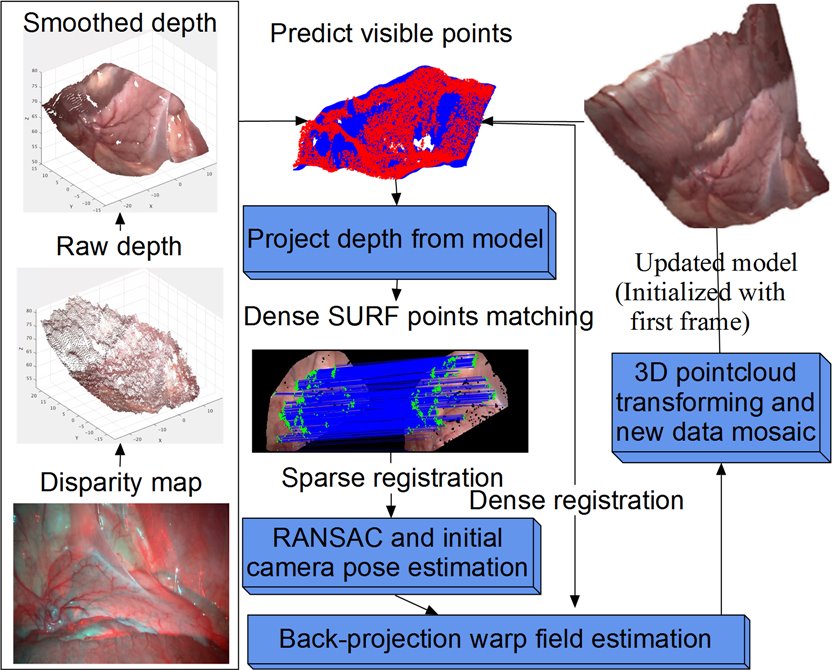}
	\caption{The framework of our deformable soft-tissue reconstruction.}
	\label{fig:pipeline}
\end{figure}

Many research activities have been devoted to deal with 3D soft-tissues reconstruction. A structure from motion pipeline \cite{lin2015video} is proposed for partial 3D surgical scene reconstruction and localization. And in \cite{stoyanov2012stereoscopic}, stereo images were used to extract sparse 3D point cloud. \cite{haouchine2015monocular} and \cite{malti2011template} extract whole tissue surface of organs from stereo or monocular images. \cite{grasa2011ekf} and \cite{lin2013simultaneous} adopt conventional Simultaneous Localization and Mapping (SLAM) approaches, extended Kalman filter (EKF) SLAM and Parallel Tracking and Mapping, with modifications on separating rigid points and non-rigid points for tracking and sparse mapping. Contrary to feature extraction based methods, Du et al. \cite{du2015robust} employed an optical flow based approach namely Deformable Lucas-Kanade for tracking tissue surface. All the methods described above contribute greatly to enable implementing augmented reality or virtual reality in computer assisted interventions which will greatly promote the accuracy and efficiency of MIS. Yet, these work mainly focus on tracking key feature points for localization and sparse mapping and no work has been devoted to geometry based registration and dense dynamic soft-tissue surface  reconstruction and dynamic deformation visualization. \par

With the development of consumer-based RGBD cameras like Microsoft Kinect and Intel Realsense, volume based template free reconstruction method has been proposed in reconstructing deformable object and mainly human body. All related works follow the basic ideas presented in Kinect Fusion \cite{newcombe2011kinectfusion} which makes use of truncated signed distance function (TSDF) for fusing and smoothing rigid objects in real-time. Inspired by this idea, research efforts are devoted to transferring TSDF fusion approach into modelling non-rigid objects. By dynamically warping TSDF volumes,  \cite{newcombe2015dynamicfusion} and \cite{innmann2016volumedeform} achieved real-time non-rigid model deformation and incremental reconstruction. These template-free techniques are able to process slow motion without occlusions because the sensor used is a single depth camera. Meanwhile, \cite{dou2016fusion4d} proposed a system consisting multiple RGBD cameras as a substitution so that a real-time colored, fast moving and close-loop model can be built. This work is mainly integrated into the Holoportation system whose robustness in handling fast movement benefits from multi-view cameras with fixed position. Although promising results can be achieved, none of these depth-sensor based approaches can be directly applied into the surgical scenario due to limitation of the sensor size and the requirement of high accuracy in the clinical application. And all the works mentioned above are fixed volume based data management approach which requires spatial limits of the scenario. Thus, none of these methods can be directly applied to the computer-assisted interventions in MIS.\par

Since point cloud can be acquired from disparity of stereo images \cite{maier2013optical},  we proposed a new framework for mapping the deformed soft-tissue using a stereoscope. There are two major requirements in clinical application which limits applying DynamicFusion \cite{newcombe2015dynamicfusion} like pipeline into surgical vision. First, due to the spatial and computational limitations, TSDF pipeline requires a predefined volume and only allows target object move within this boundary. While in the MIS scenario, due to the limited field of view of the scope, the surgeons always require the scope moves freely in the space in order to observe more areas of the tissue during interventions. Volume deforming approach used in \cite{newcombe2015dynamicfusion}, \cite{innmann2016volumedeform} and \cite{dou2016fusion4d} makes computation and unnecessary data storage increases exponentially as the volume size increases and there is a trade-off between model details (depending on the grid size) and computational cost in volume based data management. Second, different from obvious topologies in dynamic human body modeling, smoothness of organs makes the algorithm easily converges to a local minimum. Considering the small field of view of scope, the drifts of reconstruction caused by mismatching can seldom be corrected. This is different from the scenario of large field of view since they can frequently re-observe the target as loop closure \cite{newcombe2015dynamicfusion} or even reset the model \cite{dou2016fusion4d} if multiple cameras are provided (8 set of depth cameras were used in \cite{dou2016fusion4d}). Even a slight drift will lead to misalignment in textures especially on vascular.
\par

In this paper, we proposed an innovative (SLAM) framework to recover the dense deformed 3D structure of the soft-tissues in MIS scenario. Here, point cloud based method is proposed instead of volume based model management which not only avoids the blurring surface but also can manage the texture/color information at the same time for model rendering and feature points extraction combined with new observation. We also applied dense Speeded Up Robust Feature (SURF) descriptors for providing mass number of pair-wise registering key points which can greatly overcome the texture gaps caused by the error of the reconstruction of the deformable tissue. This error is due to the smoothness of the surface, which has been addressed in case of clothes \cite{innmann2016volumedeform}. In all, this is the first paper in MIS community that can dynamically reconstruct deformable dense RGB model in near real-time. \par

This paper is organized as follows: Section II describes the technical details of model management, deformation field estimation and the approach of fusing new observations. Section III shows the results to validate the proposed algorithm with some discussions of the pros and cons. Section IV concludes the paper with some prospects of future work.

\section{Methodology}

\subsection{Framework overview}

Our framework for recovering and fusing the deformation of the soft-tissue consists of depth estimation from stereo images, sparse key points extraction and matching, warping field estimation and new data fusion (see Fig. \ref{fig:pipeline}). The model (point cloud with RGB information) will be initialized by the first estimated depth. Then, each time when new depth is incorporated, visible points will be estimated and dense and sparse registration will be processed to estimate optimum warping field. After that, the model will be deformed and fused with new observations. Sections B-G will explain all the modules in details.\par

Different from \cite{newcombe2015dynamicfusion}, \cite{innmann2016volumedeform} and \cite{dou2016fusion4d} which use 3D volume named TSDF as model management tool and extract mesh (structured 3D surface with vertices and triangle faces) for estimating warping field, we directly acquire dense point cloud as data management. Each point stores 3 properties: coordinate $\bm{v}_i$, weight $\omega(\bm{v}_i)$ and color $\bm{C_i}$. The difference in data management affects our pipeline fundamentally. First, previous approaches applied marching cube to extract mesh from the 3D volume at each frame due to the requirement of estimating the visible points in the warping field. While, the point cloud data management proposed in this paper does not require any marching cube points extraction process, and efficient real-time live model rendering can be easily achieved. Besides, data management in our pipeline enables model to move freely without the predefined range in volume based method. As a matter of fact, in clinic it is annoying or even impossible for surgeons to predefine the volume range. In this way, surgeons can perform the reconstruction at any time without pre-requisite range and grid size setting steps. Another important benefit is that with point cloud data management, the estimation of the warping field at the current frame is from the warping field estimated at the previous frame instead of from the model at the initial frame as in \cite{newcombe2015dynamicfusion} and \cite{innmann2016volumedeform}. A canonical model, or model in the initial frame, is not required at all in the proposed framework. Using the method proposed in this paper, the latest observation can be registered consecutively to the previous model instead of the initial model, which results in a more accurate estimation of the warping field. In fact, \cite{dou2016fusion4d} periodically reset the entire volume to handle large misalignments caused by deforming the initial model to last state that cannot be recovered from the per-voxel refresh.\par

In the proposed framework, first, depth is estimated from the stereo RGB images captured from the scope intra-operatively. And the model is initialized by the point cloud with colors from the first frame. Then, at each time new stereo images are acquired, we extract potential visible points from the model and project them into 2D RGB images. Dense SURF algorithm is applied to extract massive number of key points for running an initial rigid point cloud transformation estimation. This sets model to a good initial position for better warping field estimation. Optimal warping field is estimated by minimizing energy function in the form of sum of squared distance between sparse key points and dense visible points. After the estimation of the warping field, we deformed the last updated model to fit with new observation, predicted the unobserved tissues by the As-Rigid-As-Possible principle \cite{sorkine2007rigid} and fused new observation in the model. Through this pipeline, live model with deformation can keep tracked with new observations and built incrementally.

\subsection{Depth estimation from stereo images}
We adopt Efficient Large-scale Stereo (ELAS) \cite{billings2012system} as the depth estimation method. Originally designed to map large scale scenario in near real time, ELAS has also been proved to achieve good result in surgical vision \cite{zhang2016autonomous}. Therefore, we applied this method as the process for providing initial colored depth for soft-tissues from stereo images.

\subsection{Sparse key correspondence and camera pose estimation}
\label{SparseKeyPointsSec}

As described in \cite{innmann2016volumedeform}, if input lacks geometric features, dense depth-based alignment is ill-posed and results in significant drifts \cite{volumedeformurl}. This is always the case in the clinic scenario because almost all the tissues are smooth and most of them lack obvious geometry information. Therefore, we applied similar strategies as in \cite{innmann2016volumedeform} by adding control anchor points to enhance the robustness. Different from \cite{innmann2016volumedeform} which applied a typical SIFT on two consecutive images as anchor points extraction method, dense SURF is used  which has been proposed to be able to get dense feature points descriptors \cite{uijlings2009real}  for model to frame registration. The basic idea of dense SURF is to directly set dense grid of locations on a fixed scale and orientation instead of detecting spatially invariant corner points. In this way, dense SURF provide much denser key points than conventional SIFT or SURF approach. The reason we use it here is to deal with motion blur and low quality images resulted from the fast movement of the scope, while the original SIFT or SURF methods cannot provide enough key points in this case. Besides, spatially scattered key points greatly enhance stability of texture in the overlapping region.	To speed up the process, we set the stride to 3 pixels. Another difference with \cite{newcombe2015dynamicfusion} is that for stability we project the colored point cloud to RGB and depth map in last camera coordinate and run dense SURF between projected RGB and depth map and new left RGB scope image. Visible points from point cloud map with RGB colors will be projected into a `model RGB image' and matched with new RGB image observation. In this way, we avoid drift issue in the model to frame approaches. After extracting dense key points, we lift them into 3D coordinate.\par

After acquiring key feature points correspondences from dense SURF, we process an iterative rigid pose estimation based on Random Sample Consensus (RANSAC) with these key points not only to estimate a rigid translation and rotation parameters which will be used as the initial guess in the optimization to estimate the warping field, but also to filter the outliers which doesn't fall into a threshold. After first estimation and outliers filtering, we get rigid pose again to gain more accurate initial pose. The estimated global parameters provide a good initial input for later deformation parameter estimation. \par

\subsection{Deformation parameters}

We employ the embedded deformation (ED) which was first proposed in \cite{sumner2007embedded} and then used in \cite{dou2016fusion4d} as a free form surface deformation approach. The warping field is made up of a set of uniformly scattered sparse ED nodes companied by an affine matrix in $\mathbb{R}^{3\times3}$ and a translation vector in $\mathbb{R}^3$. Each source point on the original model will be transformed to the target position by several nearest ED nodes and the influence their exertion depends on the distance to the ED nodes. \par
The $j$th ED node is described by a position $\bm{g_j}$ $\in\mathbb{R}^3$, a corresponding quasi rotation (affine) matrix $\bm{A_j}$ $\in\mathbb{R}^{3\times3}$ and a translation vector $\bm{t_j}$ $\in\mathbb{R}^3$. For any given point $\bm{p}$, it is mapped to a new locally deformed point $\bm{\tilde{p}}$ by the ED warping field.  \par
\begin{equation}
\bm{\tilde{p}}=\bm{A_j}(\bm{p}-\bm{g_j})+\bm{g_j}+\bm{t_j}
\end{equation}
This non-rigid transformation can be extended to any vertex or point mapped by $k$ neighboring nodes: \par
\begin{equation}
\bm{\tilde{\bm{v}}_i}=\sum_{j=1}^k w_j(\bm{\bm{v}_i})[\bm{A_j}(\bm{\bm{v}_i}-\bm{g_j})+\bm{g_j}+\bm{t_j}]
\label{TransformationFomulation}
\end{equation}
where $w_j(\bm{\bm{v}_i})$ is quantified weight for transforming $\bm{\bm{v}_i}$ exerted by each related ED node. We confine the number of nearest nodes by defining the weight in Eq. \ref{eq_weight}. Deformation of each point in the space is limited locally by setting the weight as:
\begin{equation}
\label{eq_weight}
w_j(\bm{\bm{v}_i})=(1-||\bm{\bm{v}_i}-\bm{g_j}||/d_{max}).
\end{equation}
where $d_{max}$ is the maximum distance of the vertex to $k + 1$ nearest ED node.

\subsection{Energy function}
The objective function formulated is composed of four terms: Rotation constraint, Regularization, the point-to-plane distances between the visible points and the target scan and sparse key points correspondence as:\par
\begin{equation}
\argmin\limits_{\bm{\bm{A_1}},\bm{\bm{t_1}}...\bm{A_m},\bm{t_m}} w_{rot}E_{rot}+w_{reg} E_{reg}+w_{data} E_{data}+w_{corr} E_{corr}
\label{energyfunction}
\end{equation}
where $m$ is the number of ED nodes. Here, all the variables in the state vector for this energy function are the $[\bm{A_j},\bm{t_j}]$ from each ED node. \par
To prevent the optimization converging to an unreasonable deformation, here we follow the method proposed in \cite{sumner2007embedded} which constrains the model with Rotation and Regularization.  \par
\textbf{Rotation} $E_{rot}$ sums the rotation error of all the matrix in the following form:
\begin{equation}
E_{rot}=\sum_{j=1}^m Rot(\bm{A_j})
\end{equation}
\begin{equation}
\begin{aligned}
Rot(\bm{A})=(\bm{c_1}\cdot\bm{c_2})^2+(\bm{c_1}\cdot\bm{c_3})^2+(\bm{c_2}\cdot\bm{c_3})^2+\\
(\bm{c_1}\cdot\bm{c_1}-1)^2+(\bm{c_2}\cdot\bm{c_2}-1)^2+(\bm{c_3}\cdot\bm{c_3}-1)^2
\end{aligned}
\end{equation}
where $\bm{c_1}$, $\bm{c_2}$ and $\bm{c_3}$ are the column vectors of the matrix $\bm{A}$.

\textbf{Regularization}. The basic idea for this term is to prevent divergence of the neighboring nodes exerts on the overlapping space. The quantity for this term represents the difference of deformation exerted by the neighboring node and itself should be almost the same. Otherwise, the surface will not be smooth. Therefore, we introduce the term $E_{reg}$ to sum the transformation errors from each ED node. Note that a huge weight of regularization makes the non-rigid transformation deteriorate to the rigid transformation.\par
\begin{equation}
E_{reg}=\sum_{j=1}^m\sum_{k\in{\mathbb{N}(j)}} \alpha_{jk}||\bm{A_j}(\bm{\bm{g_k}}-\bm{g_j})+\bm{g_j}+\bm{t_j}-(\bm{\bm{g_k}}+\bm{t_k})||^2
\end{equation}
where $\alpha_{jk}$ is the weight calculated by the Euclidean distance of the two ED nodes. In \cite{sumner2007embedded}, $\alpha_{jk}$ is uniformly set to 1. $\mathbb{N}(j)$ is the set of all neighboring nodes to the node $j$. \par
\noindent\textbf{Data Term}. 
We first build a distance field volume with voxels recording distances to nearest depth points. Then traverse all the model points and extract visible points which falls into a threshold. These visible points $\bm{v}_i ~(i\in\{1,...,N\})$ are considered to be visible and applied in object function for optimization. \par
Similar to \cite{newcombe2015dynamicfusion}, \cite{innmann2016volumedeform} and \cite{dou2016fusion4d}, we adopt back-projection approach as a practical model registration strategy that penalizes misalignment of the predicted visible points $\bm{v}_i ~(i\in\{1,...,N\})$ and current depth scan $\mathbb{D}$. Data Term is sum of point-to-plane errors in the form of: \par
\begin{equation}
E_{data}=\sum_{i=1}^N (\bm{\textbf{n}}(P(\tilde{\bm{v}_i}))^T(\bm{\tilde{\bm{v}}_i}-\Gamma(P(\tilde{\bm{v}_i}))))^2
\end{equation}
where $\Gamma(\bm{v}) = \Pi(P(\bm{v}))$ and $\tilde{\bm{\textbf{n}}}(u)$ is the corresponding normal to the pixel $u$ in the depth $\mathbb{D}(u)$. $\bm{\tilde{\bm{v}}_i}$ is the transformed position of point $\bm{v}_i$. $P(\bm{v})$ is the projective ($\mathbb{R}^3 \rightarrow \mathbb{R}^2$) function for projecting visible points to depth image. $\Pi(u)$ is the back-projection function for lifting the corresponding pixel $P(\bm{v})$ back into 3D space ($\mathbb{R}^2 \rightarrow \mathbb{R}^3$). \par
Similar to \cite{newcombe2015dynamicfusion}\cite{innmann2016volumedeform}\cite{dou2016fusion4d}, we adopt the strategy for extracting visible points from last state of model by filtering with the threshold of distance and normal angles to current depth. Where $\epsilon_d$ and $\epsilon_n$ are thresholds. \par 

\begin{equation}
||\bm{\bm{v}}_i-\Gamma(P(\bm{v}_i))|| < \epsilon_d,  \quad
\bm{\textbf{n}}(\bm{v}_i) \cdot \bm{\textbf{n}}(P(\bm{v}_i)) < \epsilon_n
\end{equation}

As described in \cite{dou2016fusion4d}, back-projection and point-to-plane strategies make full use of the input depth image so that the Jacobians can be calculated in 2D which leads to fast convergence and robustness to outliers. As depth generated from stereo images are not as accurate as that from depth sensors like Kinect, the visual hull terms recommended by \cite{dou2016fusion4d} is not applied because the empty space and free space are not actually observed due to the misalignment of disparity maps. \par
\noindent\textbf{Correspondence Term} is measured by the Euclidean distance between pair-wise sparse key points generated from dense SURF described in Section  \ref{SparseKeyPointsSec} in the following form:
\begin{equation}
E_{corr}=||\bm{\tilde{\bm{V}}_i-\bm{V}_i||}
\end{equation}
where $\tilde{\bm{V}}_i$ and $\bm{V}_i$ are the new and old position of control points.
\subsection{Optimization}
Both registration and deformation parameter estimation processes are carried out simultaneously by minimizing the energy functions. Here, we use Levenberg-Marquardt (LM) algorithm to solve the nonlinear optimization problem.\par 

Different from conventional Gauss-Newton (GN) optimization method, LM introduces an extra term $\mu\bm{I}$ ($\bm{I}$ is the identity matrix and $\mu$ is a ratio) which controls the aggressiveness of GN approach. If the step is over confident, the damping factor $\mu$ will be increased. Otherwise it will be decreased. Another key point is that solving global and local transformation together will lead to singularity in solving the linear equation which is caused by the fact that ED parameters also contain information about global rotation and transformation. Through LM algorithm, this numerical problem can be prevented.\par

\subsection{Model update by new observation}
Previous works adopt TSDF volume to store and fuse models \cite{newcombe2011kinectfusion} \cite{newcombe2015dynamicfusion} \cite{innmann2016volumedeform} \cite{dou2016fusion4d}. Fine mesh can be generated in real-time by marching cube method. Nevertheless, all these volume based
approaches are unable to work in surgical vision because of the unknown spatial range of the target soft-tissues. To overcome this restriction, we proposed a weighted point cloud to represent the built model. In our algorithm, each point records exactly the surface location with weight showing how certain it believes the record.  \par

After acquiring an appropriate warping field, we transform all the points to their deformed positions and predict visible points again. For each updated point, a truncated signed distance weight (TSDW) is assigned to each pixel of new depth: 
\begin{equation}
\omega(\bm{p}_n) =
\begin{cases}
d_{min}(\bm{p}_n) / (0.5 * \epsilon) &\mbox{if $abs(\bm{\bm{\tilde{v}}_i}|_z - \mathbb{D}(P(\bm{p}_n)))<\tau$}\\
0 &\mbox{otherwise}
\end{cases}
\end{equation}
\begin{figure}[]		
	\centering
	\subfloat[dense SURF]{
		\label{fig:scanmatching_a}
		\begin{minipage}[htpb]{0.4\textwidth}
			\centering
			\includegraphics[width=1\linewidth]{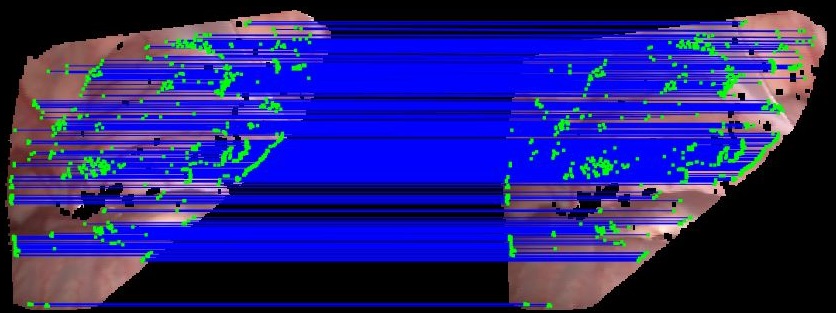}
		\end{minipage}
	}%
	\\
	\subfloat[SIFT]{
		\label{fig:scanmatching_c}
		\begin{minipage}[htpb]{0.4\textwidth}
			\centering
			\includegraphics[width=1\linewidth]{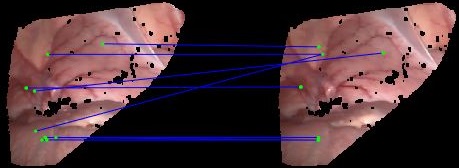}
		\end{minipage}
	}
	\caption{The comparison between the dense SURF and SIFT using stereo videos of abdomen wall. Results imply that dense SURF can generate more key points which are critical in soft-tissue matching while SIFT produce less or even no correspondences.}
	\label{fig:3matching comparisons}
\end{figure}
\captionsetup[subfigure]{labelformat=empty}
\begin{figure}[]	
		\centering
		\subfloat[]{
			\label{fig:scanmatching_a}
			\begin{minipage}[htpb]{0.45\textwidth}
				\centering
				\includegraphics[width=0.9\linewidth]{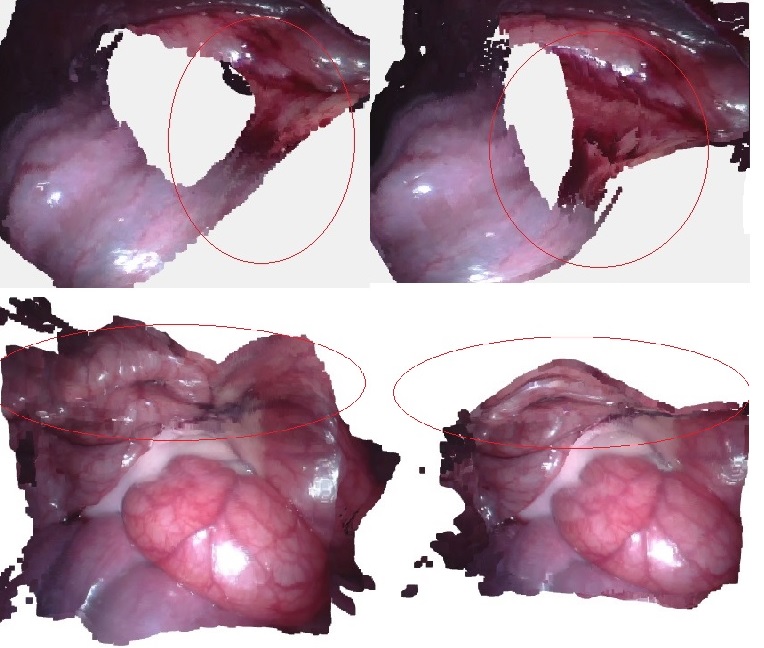}
			\end{minipage}
		}\/	
	\caption{The comparison between pipeline with and without dense SURF constraint (Left is with constraint while right is without). Significant errors happen either in texture or in topologies without SURF constraint.}
	\label{fig:KeyPointsCompare}
\end{figure}

\setlength{\parindent}{0pt}where $d_{min}(\bm{p})$ is the minimum distance of point $\bm{p}$ to its corresponding nodes and $\epsilon$ is the average grid size of nodes. We discard the $z$ directional difference if it is too large because this is probably due to inaccurate warp field estimation. We fuse current depth generated from model with new depth by:
\begin{equation}
\mathbb{D}_{n+1}(P(\bm{p}_n)) = \frac{\tilde{\bm{v}}_i|_z \omega(\bm{p}_{n-1})+\mathbb{D}_n(P(\bm{p}_n))}{\omega(\bm{p}_{n-1})+1}
\end{equation}
\begin{equation}
\omega(\bm{p}_n)=min(\omega(\bm{p}_{n-1})+1 ,\omega_{max})
\end{equation}
where $\tilde{\bm{v}}_i|_z$ is the value of point $\bm{v}_i$ on the z direction and $\bm{p}$ is the projected pixel $P(\tilde{\bm{v}}_i)$ of the transformed points $\bm{v}_i$. Different from rigid transformation where uncertainty of all the points in 3D space are considered as equal, in the case of non-rigid fusion, if a point is further away to the nodes of warping field, we are less likely to believe the registered depth \cite{newcombe2015dynamicfusion}. Therefore, we practically measure this certainty by using the minimum distance from point to nodes and regularize it with half of the unified node distance. The upper bound of weight is set to 10. \par
Therefore, by a correct warping field, we can perform almost the same fusion process as what TSDF does, but do not require a predetermined volume. Weighted points based method offers a variety of benefits: (1) Points based data management and fusion free the geometry extent while still maintain the ability of begetting a smooth fused surface. (2) By using the points based management, all the components in the proposed framework, e.g. visible points prediction, warping field estimation and model update are unified in points. Process like conversion between volume and mesh are not necessary anymore. (3) Our approach enables current live model updated from the model in the previous step instead of from the reference model which prevents drifts accumulated in the warping field. \par
\captionsetup[subfigure]{labelformat=empty}
\vspace{0pt} 
\begin{figure*}[h]	

	\centering
	\subfloat[Frame 1]{	
		\begin{minipage}[htpb]{0.17\textwidth}	
			\centering
			\includegraphics[width=1\linewidth]{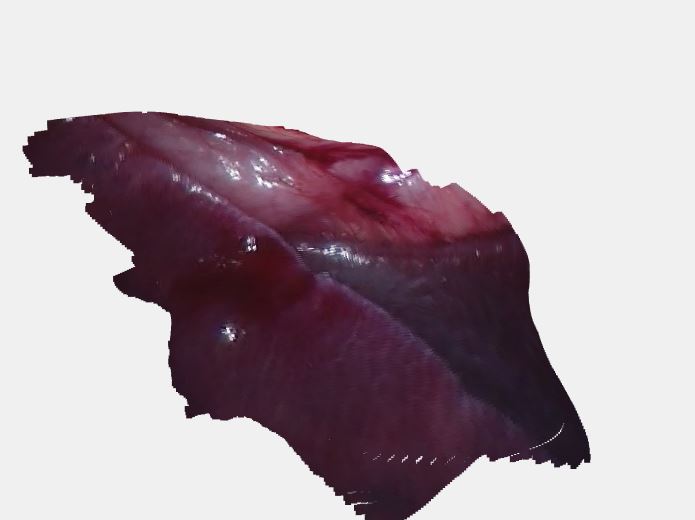}			
		\end{minipage}				
	}%
	\subfloat[Frame 30]{
		\centering
		\label{fig:scanmatching_b}
		\begin{minipage}[htpb]{0.17\textwidth}
			\centering
			\includegraphics[width=1\linewidth]{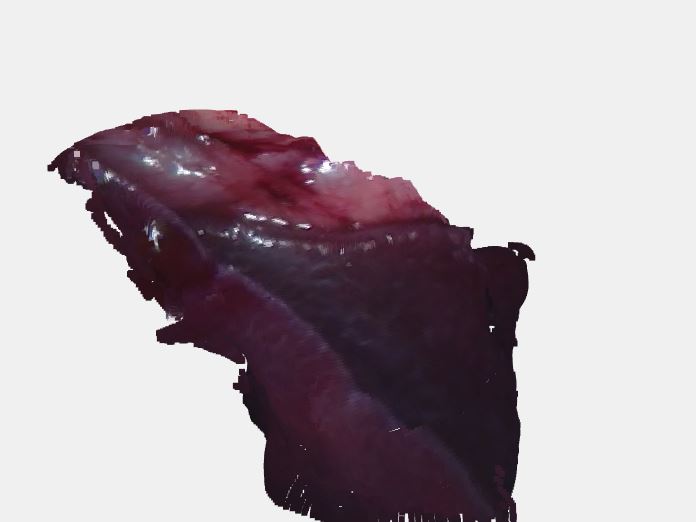}
		\end{minipage}
	}
	\subfloat[Frame 100]{
		\label{fig:scanmatching_c}
		\begin{minipage}[htpb]{0.17\textwidth}
			\centering
			\includegraphics[width=1\linewidth]{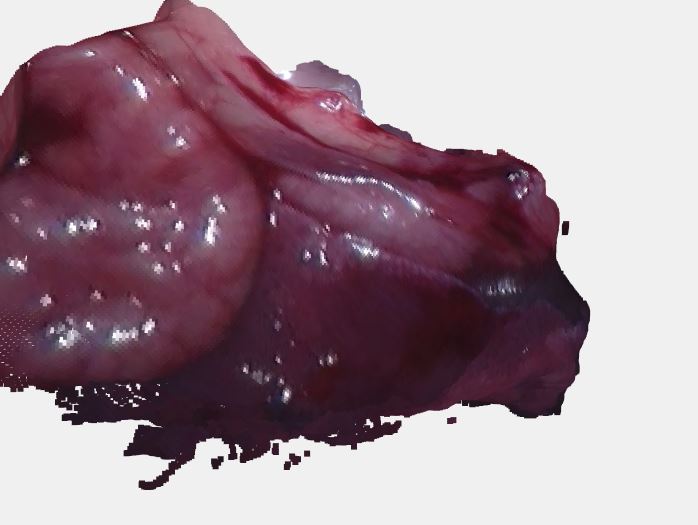}
		\end{minipage}
	}
	\subfloat[Frame 300]{
		\centering
		\label{fig:scanmatching_b}
		\begin{minipage}[htpb]{0.17\textwidth}
			\centering
			\includegraphics[width=1\linewidth]{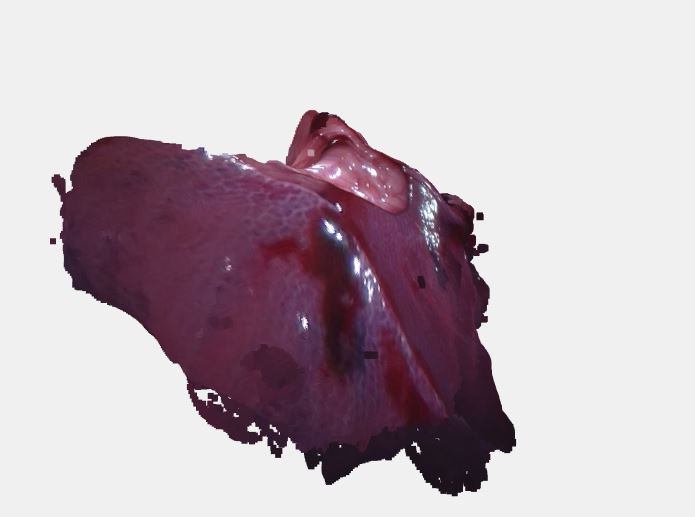}
		\end{minipage}
	}
	\subfloat[Frame 500]{
		\centering
		\label{fig:scanmatching_b}
		\begin{minipage}[htpb]{0.17\textwidth}
			\centering
			\includegraphics[width=1\linewidth]{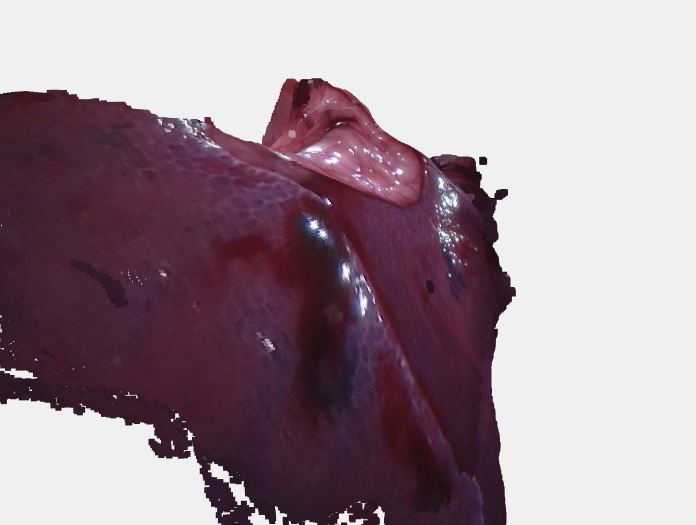}
		\end{minipage}
	}\/	
	\subfloat[Frame 1]{	
		\label{}
		\begin{minipage}[htpb]{0.17\textwidth}	
			\centering
			\includegraphics[width=1\linewidth]{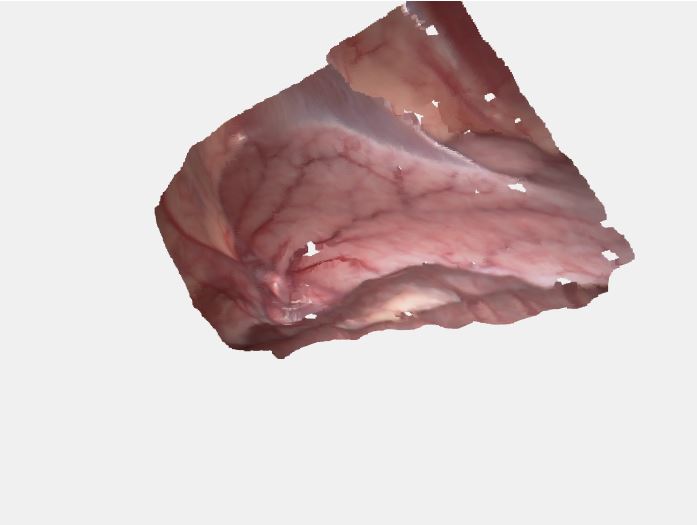}			
		\end{minipage}				
	}%
	\subfloat[Frame 30]{
		\centering
		\label{fig:scanmatching_b}
		\begin{minipage}[htpb]{0.17\textwidth}
			\centering
			\includegraphics[width=1\linewidth]{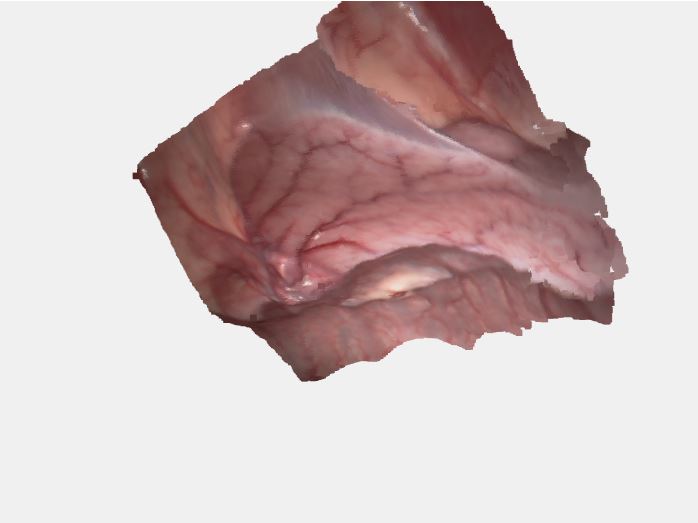}
		\end{minipage}
	}
	\subfloat[Frame 100]{
		\label{fig:scanmatching_c}
		\begin{minipage}[htpb]{0.17\textwidth}
			\centering
			\includegraphics[width=1\linewidth]{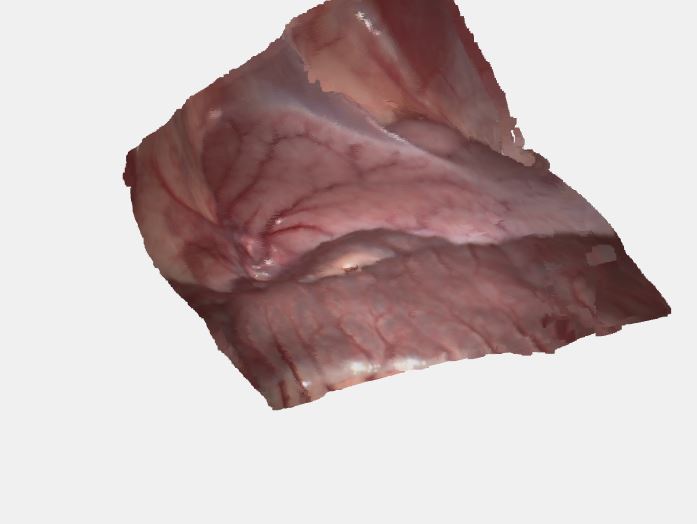}
		\end{minipage}
	}
	\subfloat[Frame 200]{
		\centering
		\label{fig:scanmatching_b}
		\begin{minipage}[htpb]{0.17\textwidth}
			\centering
			\includegraphics[width=1\linewidth]{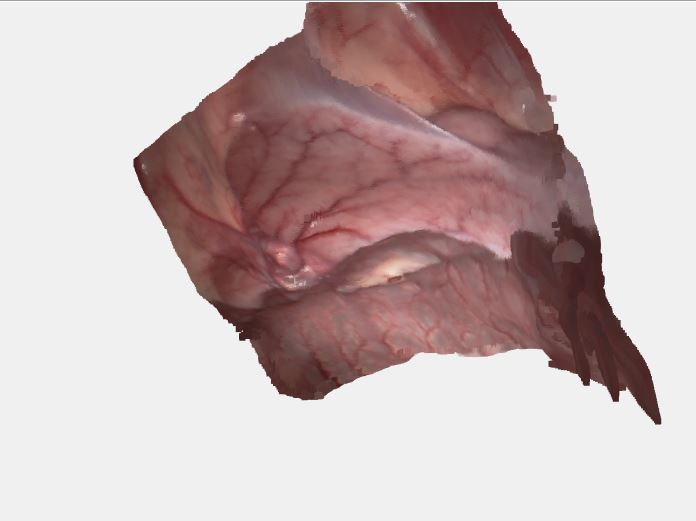}
		\end{minipage}
	}
	\subfloat[Frame 300]{
		\centering
		\label{fig:scanmatching_b}
		\begin{minipage}[htpb]{0.17\textwidth}
			\centering
			\includegraphics[width=1\linewidth]{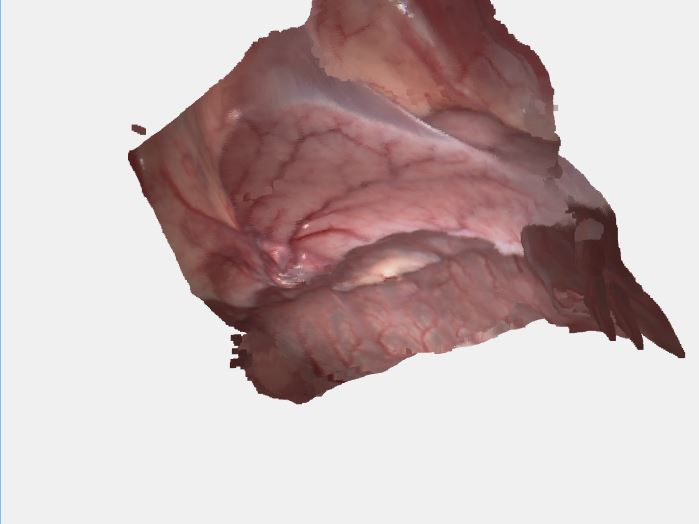}
		\end{minipage}
	}\/
	\subfloat[Frame 1]{
		\label{fig:scanmatching_a}
		\begin{minipage}[htpb]{0.17\textwidth}
			\centering
			\includegraphics[width=1\linewidth]{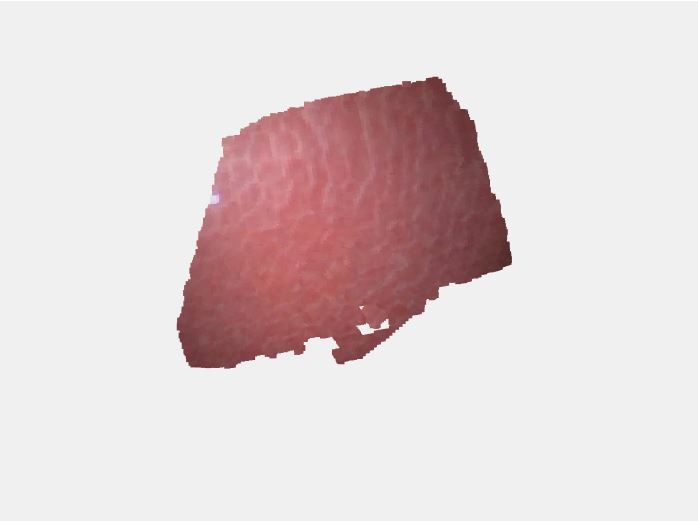}
		\end{minipage}
	}%
	\subfloat[Frame 30]{
		\centering
		\label{fig:scanmatching_b}
		\begin{minipage}[htpb]{0.17\textwidth}
			\centering
			\includegraphics[width=1\linewidth]{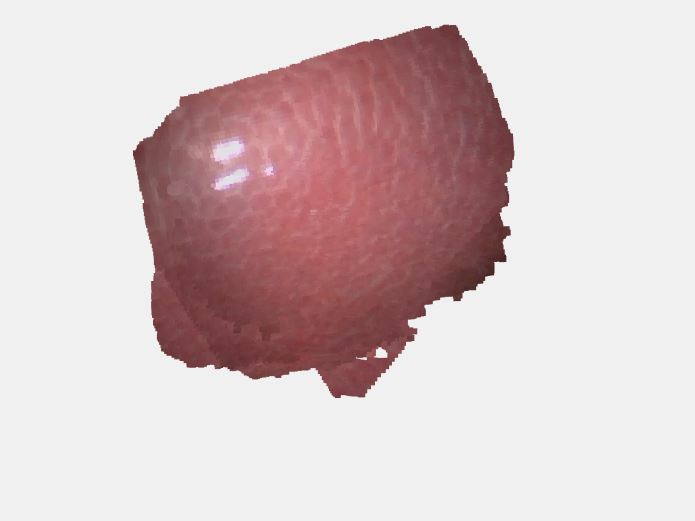}
		\end{minipage}
	}
	\subfloat[Frame 80]{
		\label{fig:scanmatching_c}
		\begin{minipage}[htpb]{0.17\textwidth}
			\centering
			\includegraphics[width=1\linewidth]{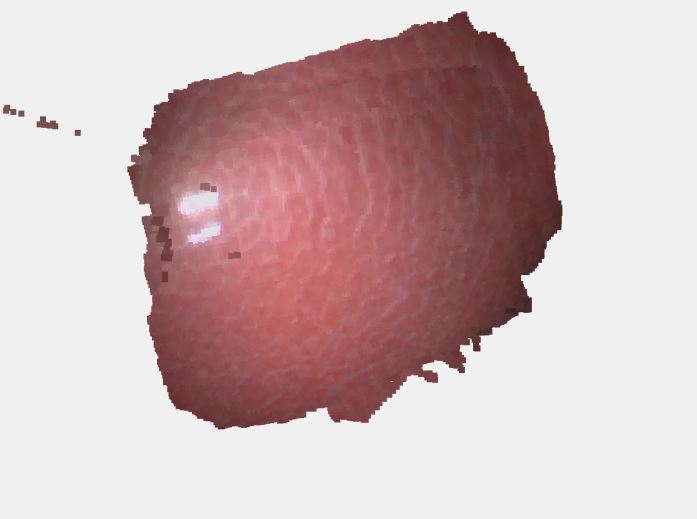}
		\end{minipage}
	}
	\subfloat[Frame 150]{
		\centering
		\label{fig:scanmatching_b}
		\begin{minipage}[htpb]{0.17\textwidth}
			\centering
			\includegraphics[width=1\linewidth]{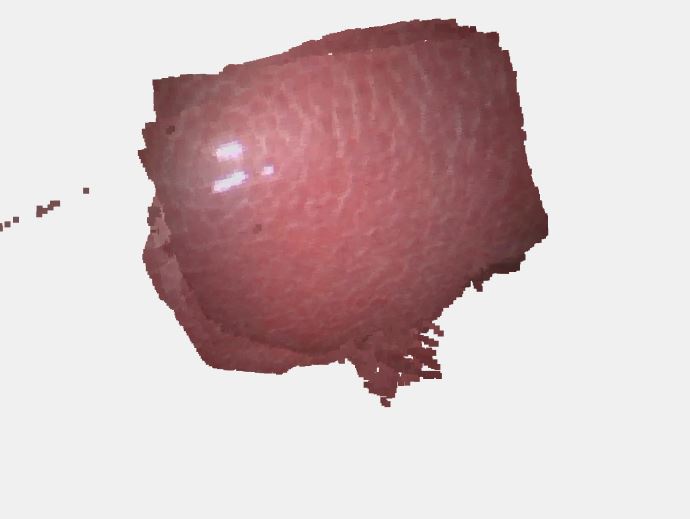}
		\end{minipage}
	}
	\subfloat[Frame 200]{
		\centering
		\label{fig:scanmatching_b}
		\begin{minipage}[htpb]{0.17\textwidth}
			\centering
			\includegraphics[width=1\linewidth]{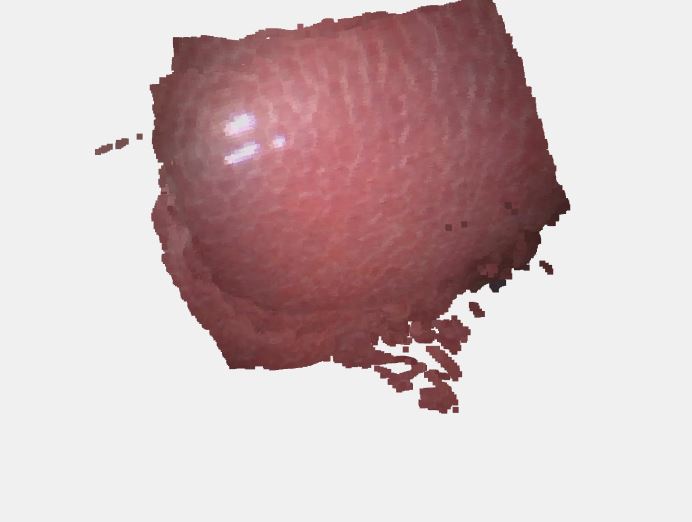}
		\end{minipage}
	}\/
	\subfloat[Frame 1]{
		\label{fig:scanmatching_a}
		\begin{minipage}[htpb]{0.17\textwidth}
			\centering
			\includegraphics[width=1\linewidth]{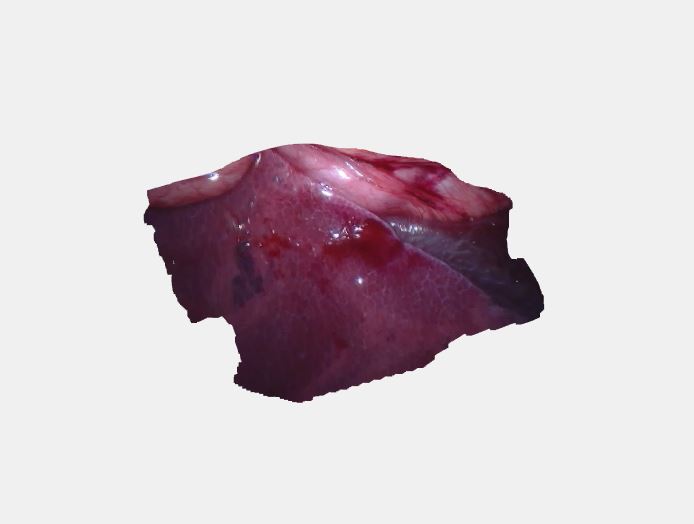}
		\end{minipage}
	}%
	\subfloat[Frame 30]{
		\centering
		\label{fig:scanmatching_b}
		\begin{minipage}[htpb]{0.17\textwidth}
			\centering
			\includegraphics[width=1\linewidth]{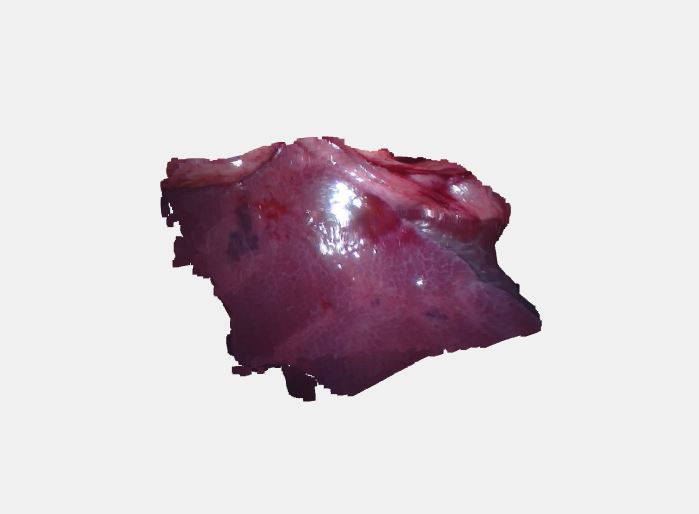}
		\end{minipage}
	}
	\subfloat[Frame 100]{
		\label{fig:scanmatching_c}
		\begin{minipage}[htpb]{0.17\textwidth}
			\centering
			\includegraphics[width=1\linewidth]{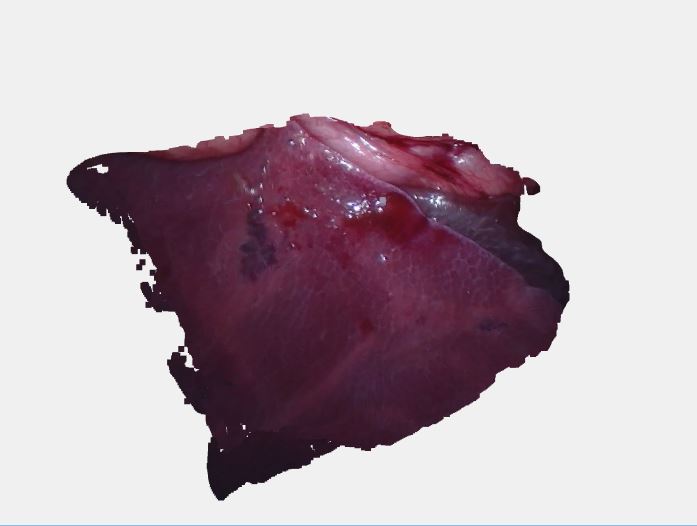}
		\end{minipage}
	}
	\subfloat[Frame 300]{
		\centering
		\label{fig:scanmatching_b}
		\begin{minipage}[htpb]{0.17\textwidth}
			\centering
			\includegraphics[width=1\linewidth]{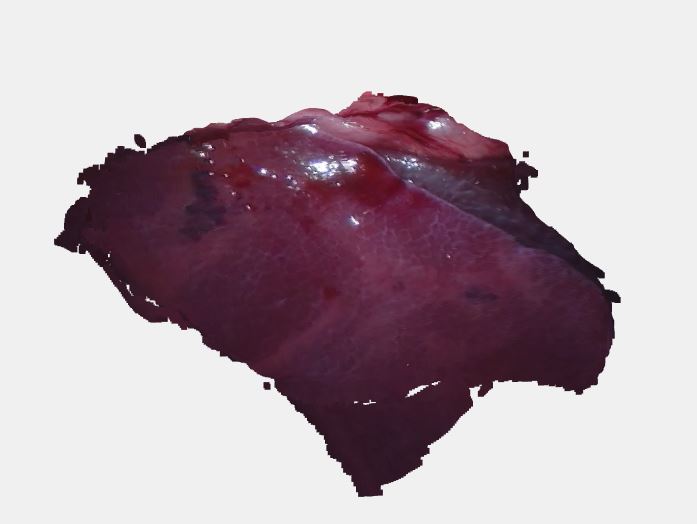}
		\end{minipage}
	}
	\subfloat[Frame 500]{
		\centering
		\label{fig:scanmatching_b}
		\begin{minipage}[htpb]{0.17\textwidth}
			\centering
			\includegraphics[width=1\linewidth]{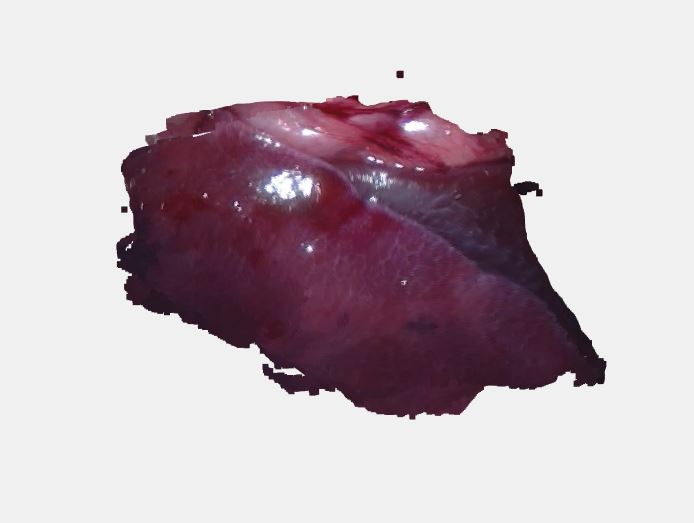}
		\end{minipage}
	}\/
	\subfloat[Frame 1]{
		\label{fig:scanmatching_a}
		\begin{minipage}[htpb]{0.17\textwidth}
			\centering
			\includegraphics[width=1\linewidth]{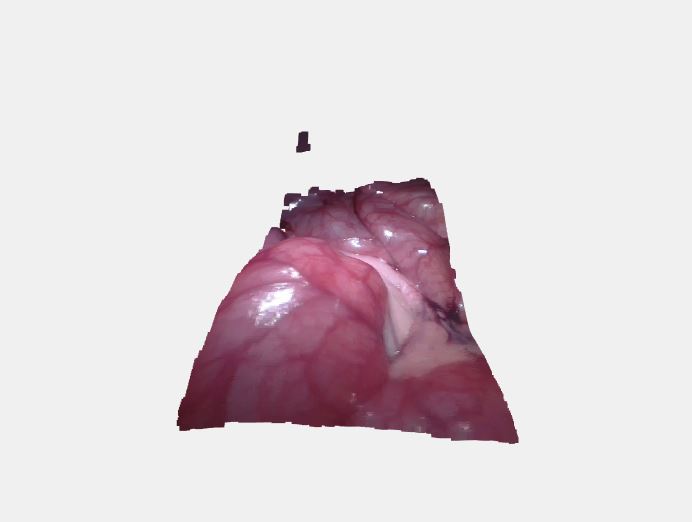}
		\end{minipage}
	}%
	\subfloat[Frame 30]{
		\centering
		\label{fig:scanmatching_b}
		\begin{minipage}[htpb]{0.17\textwidth}
			\centering
			\includegraphics[width=1\linewidth]{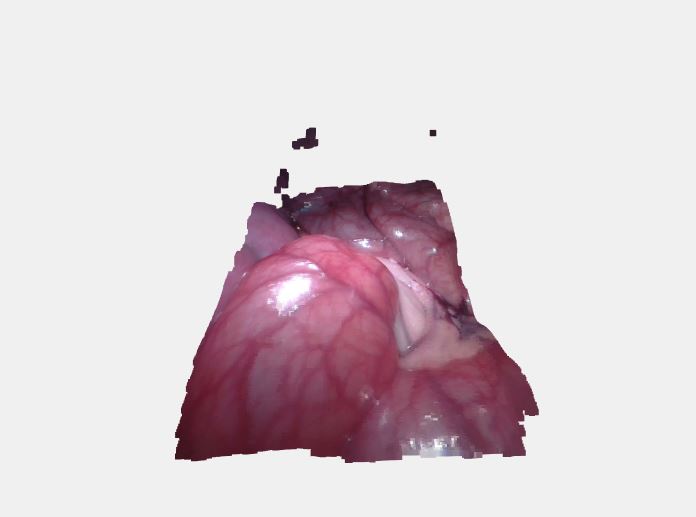}
		\end{minipage}
	}
	\subfloat[Frame 100]{
		\label{fig:scanmatching_c}
		\begin{minipage}[htpb]{0.17\textwidth}
			\centering
			\includegraphics[width=1\linewidth]{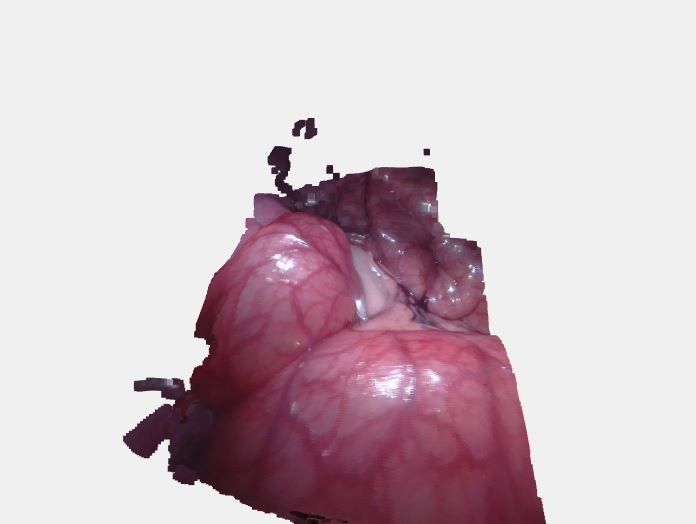}
		\end{minipage}
	}
	\subfloat[Frame 300]{
		\centering
		\label{fig:scanmatching_b}
		\begin{minipage}[htpb]{0.17\textwidth}
			\centering
			\includegraphics[width=1\linewidth]{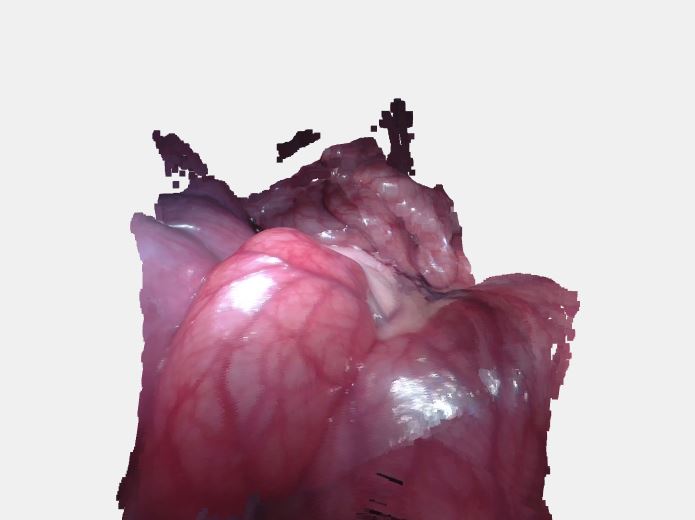}
		\end{minipage}
	}
	\subfloat[Frame 500]{
		\centering
		\label{fig:scanmatching_b}
		\begin{minipage}[htpb]{0.17\textwidth}
			\centering
			\includegraphics[width=1\linewidth]{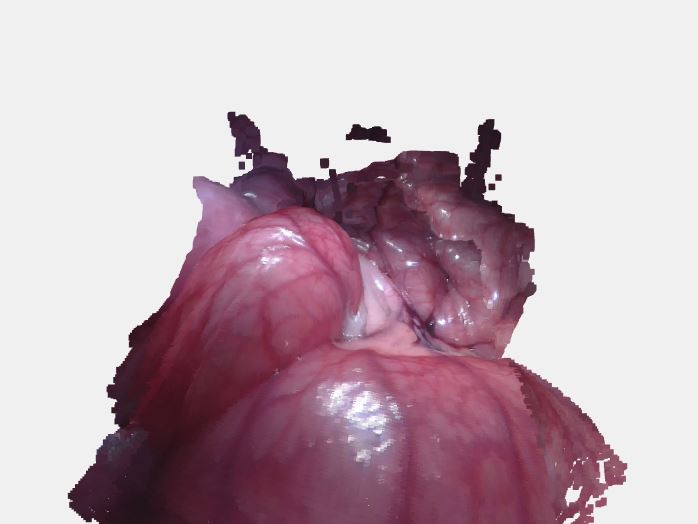}
		\end{minipage}
	}\/
	\caption{ Non-rigid reconstruction of different soft tissues using in-vivo datasets. Pictures present the whole constructed model at different frames. The five videos are (from top to bottom): abdomen example1, abdomen wall, liver, abdomen example2 and abdomen example3. }
	\label{fig:5model comparisons}
\end{figure*}

\section{Results and discussion}
The SLAM framework proposed is validated using the in-vivo stereo video datasets provided by the Hamlyn Centre for Robotic Surgery \cite{giannarou2013probabilistic}. No extra sensing data other than stereo videos from the laparoscope is utilized in the process. The frame rate and size of in-vivo porcine dataset (model 2 in Fig. \ref{fig:5model comparisons}) is 30 frame per second and $640 \times 480$ while the rest is 25 frame per second and $360 \times 288$. Distance from camera to surface of soft-tissue is between 40 to 70 mm.  Due to poor quality of obtained images and some extreme fast movement of the camera, we deliberately choose the videos tested on porcine with relative slow motion and some deformation caused by respiration and tissue-tool interaction. \par

In comparison with conventional scenarios in \cite{newcombe2015dynamicfusion} \cite{innmann2016volumedeform}  \cite{dou2016fusion4d}, we conclude that there are three major challenges in surgical vision: First, scope has very narrow field of view which makes the observed information in each frame limited. Second, most of the soft-tissues have smooth surface and do not have many distinct geometric features can be applied in the registration process. In practice, we found the registration without key points results in great drifts. Last, since scope has small field of view, we encounter blurry images in the process of key points extraction.\par

\subsection{Dense SURF key points estimation}

We employ the key points strategy proposed by \cite{innmann2016volumedeform} to overcome the drift caused by smooth surface and the error of the warping field estimation. To deal with problems resulted from low quality of video and faster moving object, we applied dense SURF to extract key points. In some extreme situations, when fast movements occur in camera, no SIFT key points correspondences can be detected. \par

We set the grid sampling size of dense SURF to 3 pixels and obtained a large amount of corresponding points. In practice we found out that the extracted dense key points range from 100 to 1000 while conventional SIFT and SURF generate points from 0 to 200. After dense SURF process we refine key points and generate rigid rotation and translation by RANSAC which is a typical strategy adopted in implementing SLAM in MIS \cite{grasa2011ekf}\cite{lin2013simultaneous}. Threshold for filtering outliers is set as 2 mm similar to \cite{lin2013simultaneous}. Fig. \ref{fig:3matching comparisons} indicates that dense SURF generates enough key points for registration. Fig. \ref{fig:KeyPointsCompare} shows that registration without key points makes the registration process converges locally which either results in disorder of texture or squeezes on the soft-tissue. \par

\begin{figure}[]		
	\captionsetup[subfigure]{labelformat=empty}
	\centering
	\subfloat[]{
		\label{fig:heart_Axial}
		\begin{minipage}[htpb]{0.4\textwidth}
			\centering
			\includegraphics[width=1\linewidth]{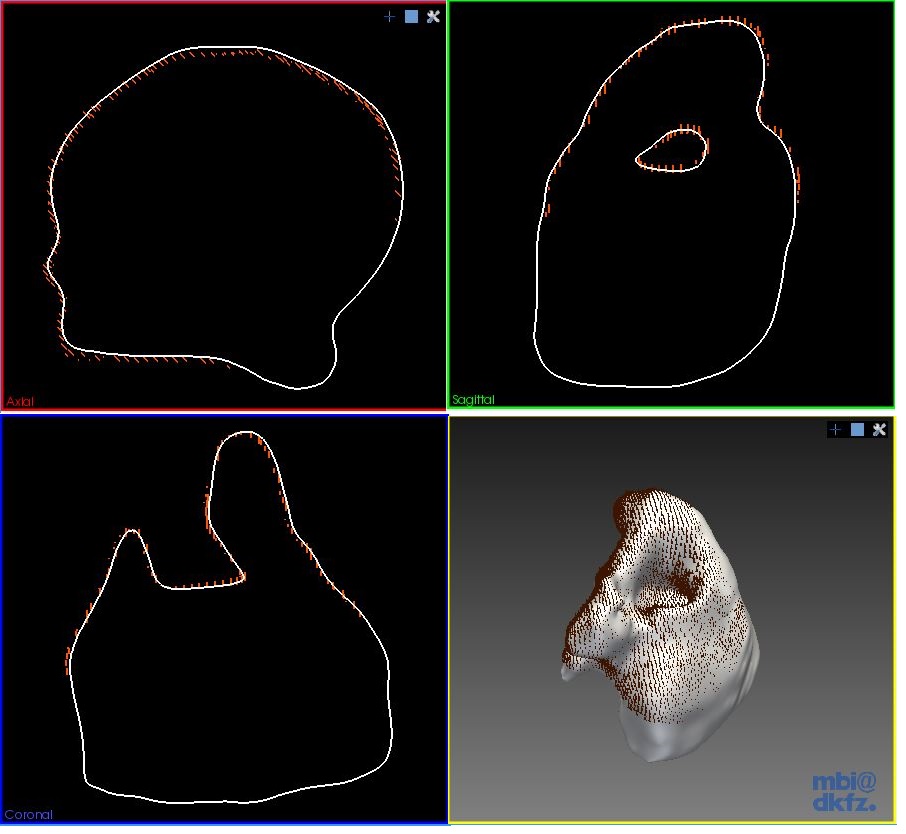}
		\end{minipage}
	}%
	\caption{The Axial, Coronal, Sagittal and 3D views of the deformed model and ground truth at the last frame (heart). The red points denote the scan of the last frame. }
	\label{fig:6traverse_map}
\end{figure}
\subsection{Deforming the model and fusing new depth}

The point cloud density is set to 0.2mm and node density is set to 4mm. Point cloud downsampling process is carried out by setting a fixed box to average points fill inside each 3-D box. The weights for optimization 
are chosen as $w_{rot}=1000, w_{reg}=10000, w_{data}=1, w_{corr}=1$. We measure the error by subtracting projected model image and the observed depth image. \par
We found out that the depths generated from fast moving camera are in low quality. Therefore, threshold is applied to discard some frames when average errors are above. Fig. \ref{fig:5model comparisons} shows the results of soft-tissue reconstruction of our proposed SLAM framework in different frames, using 5 in-vivo laparoscope datasets \cite{giannarou2013probabilistic}. From the results we can see, the soft-tissues were reconstructed incrementally with texture. \par 
The average distance of back-projection registration of the five scenarios are 0.19mm (abdomen wall), 0.08mm (Liver), 0.21mm (Abdomen1), 0.15mm (Abdomen2) and 0.14mm (Abdomen3).\par

\subsection{GPU implementation and computational cost}\par
By parallelizing the proposed methods for General-purpose computing on graphics processing units (GPGPU), the proposed algorithm is currently implemented in CUDA with the harware ``Nvidia GTX 1060". To minimize the time cost of copying variables between GPU and CPU memories, we implement the whole approach on GPU. Current processing rate for each sample dataset is around 0.07s per frame.

\subsection{Validation using simulation and ex-vivo experiments}\par
The feasibility of our approach is validated based on simulation and ex-vivo experiment. In simulation validation process, three different soft-tissue models (heart, liver and right kidney) were downloaded from OpenHELP \cite{kenngott2015openhelp}, which were segmented from a CT scan of a healthy, young male undergoing shock room diagnostics. The deformation of the soft-tissue is simulated by randomly exerting 2-3 mm movement on a random vertex on the model with respect to the status of the deformed model from the last frame \cite{song20163d}. We randomly pick up points in the model as the accuracy is measured by averaging all the distances from the source points to target points. Fig. \ref{fig:6traverse_map} shows the final result of the simulation presented in the form of axial, coronal, sagittal and 3D maps (heart). The average error of three models are: 0.46 (heart), 0.68 (liver), 0.82  (right kidney) mm respectively. \par
\begin{figure}[]
	\captionsetup[subfigure]{labelformat=empty}
	\centering
	\includegraphics[width=0.8\linewidth]{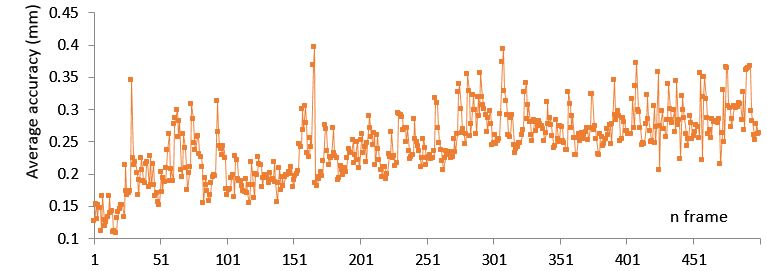}
	\centering
	\includegraphics[width=1\linewidth]{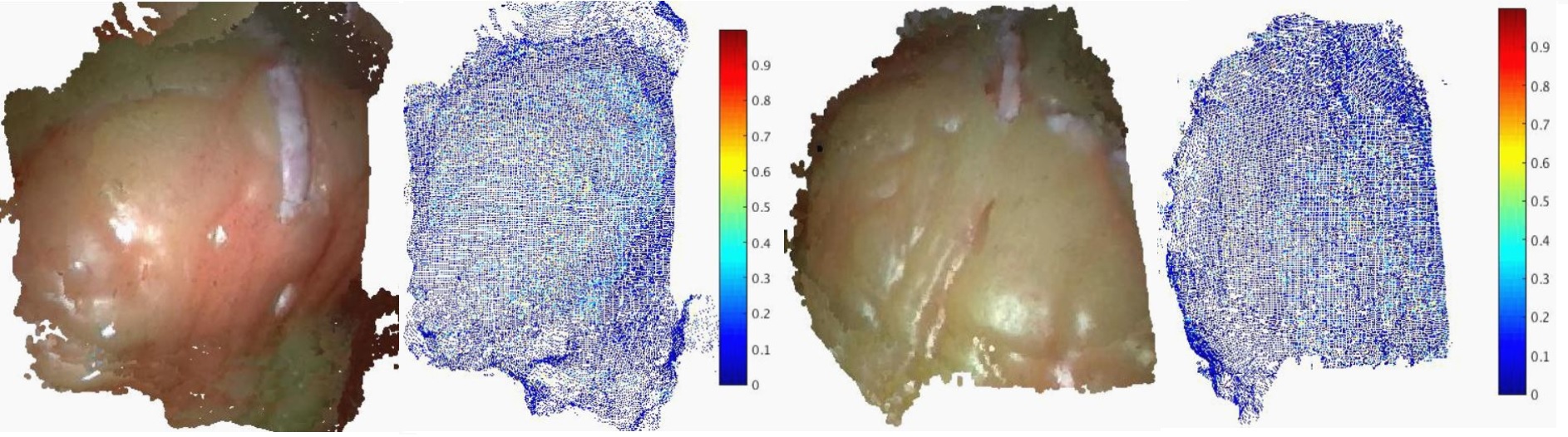}
	\caption{Ex-vivo validation with the two Hamlyn validation datasets: Silicon heart phantoms deforming with cardiac motion and associated CT scans. The upper figure is the time series of average error. The lower figures are the reconstructed geometry and corresponding error maps measured by distance to ground truth.}
	\label{fig:hamlynvalidation}
\end{figure}

We also tested the proposed approach on two ex-vivo phantom based validation dataset from Hamlyn \cite{giannarou2013probabilistic} (Fig. \ref{fig:hamlynvalidation}). As the phantom deforms periodically, we 
process the whole video and compare it with the ground truth generated from CT scan. The average accuracy of our work are 0.28mm and 0.35mm. The good result mainly comes from the abundant textures 
benefited both depth estimation and dense SURF key points extraction.\par
\subsection{Compare with VolumeDeform}
\begin{figure}[]	
	\captionsetup[subfigure]{labelformat=empty}	
	\centering
	\subfloat[]{
		\label{fig:heart_Axial}
		\begin{minipage}[htpb]{0.20\textwidth}
			\centering
			\includegraphics[width=1\linewidth]{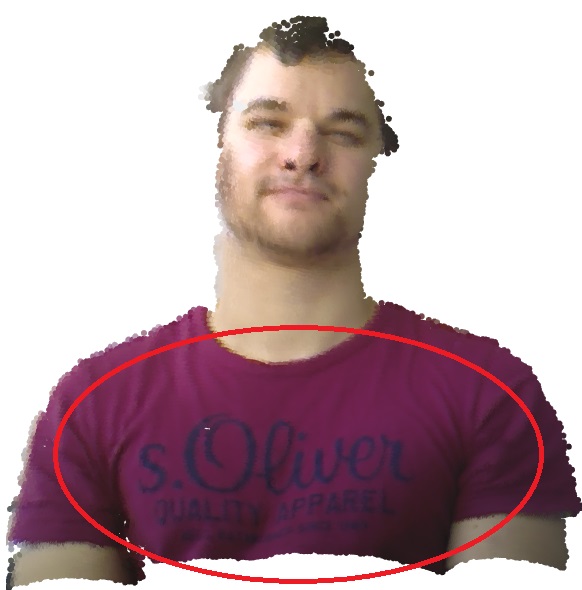}
		\end{minipage}
	}%
	\subfloat[]{
		\centering
		\label{fig:heart_Coronal}
		\begin{minipage}[htpb]{0.20\textwidth}
			\centering
			\includegraphics[width=1\linewidth]{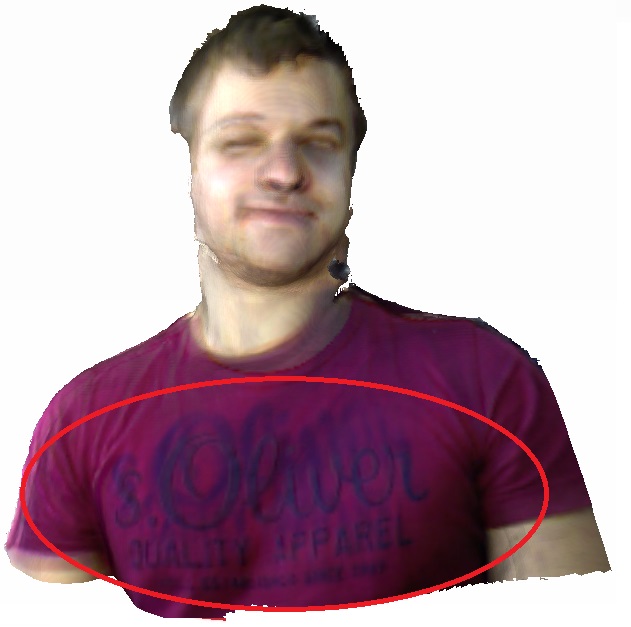}
		\end{minipage}
	}
	\caption{Comparison with VolumeDeform approach. The left is our reconstruction.The right is VolumeDeform's result. Note the difference in texture (letters on the T-shirt).}
	\label{fig:7volumedeformcompare}
\end{figure}

We also compare our work with VolumeDeform \cite{innmann2016volumedeform} by implementing a sample dataset and present the result in Fig. \ref{fig:7volumedeformcompare}. 
The source code of dynamic fusion and VolumeDeform are not available, one dataset published by \cite{innmann2016volumedeform} is used for the comparison. VolumeDeform claims less drift than DynamicFusion \cite{newcombe2015dynamicfusion} approach while our approach ensures less drift 
than VolumeDeform due to our model to frame rather than frame to frame strategy. Textures from our framework is better.\par 

\subsection{Limitations and discussions}\par
In 
the attached video, there are some illumination differences in the texture and the rapid fluctuations on the edges of the reconstructed model. The texture difference is due to different angle of light source in different image time. As we try to preserve the latest texture, we update the texture directly instead of implementing weighted average process. The rapid fluctuations on the edges results from quick wave of depth generated from ELAS. Our pipeline deforms reconstructed model to match the depth and fluctuations on the depth's edge forces model deform accordingly. This can be solved by developing a more robust depth estimation algorithm or erode the edges of the depth model. \par
While our pipeline works well on the test datasets, we would like to address the challenges facing reconstruction problem using stereoscope. The first and most important challenge is the fast movement of scope. The proposed algorithm fails to track motion when camera moving fast. Similar to traditional SLAM approaches \cite{grasa2011ekf} \cite{lin2013simultaneous}, serious consequences of fast motion are the blurry images and relevant disorder of depths. These phenomenons happen especially when current constructed model deforms to match the depth with false edges suffering from image blurring. That's why our pipeline visualize periodic deformation like respiration and heart beat clearly on central regions but shows obvious drifts on the edges. Fast motion is a challenging issue as the only data source is the blurry images. Another issue is the accuracy of texture. Laparoscope with narrow field of view results in obvious drifts and gaps on texture especially in blurry images. In the future we would like to use some image enhancement techniques to increase robustness and accuracy of our pipeline.\par

\section{Conclusion}
We proposed a dynamic deformation recovery SLAM framework for reconstructing the 3D shape of deformable soft-tissues in the scenario of MIS based stereoscope. In contrast to conventional non-rigid scene reconstruction, we replaced current volume based approach with point cloud and adjusted the fusion process for the purpose of relative large spatial requirement. Simulation and in-vivo experiments validate the feasibility of our dynamic SLAM framework. Future work will be concentrated on exploring a more robust key points extraction algorithm for enhancing robustness in situation when camera moves fast. We will also explore the feasibility of applying our method on depth generated from a monocular scope with approaches like shape from shading.

\addtolength{\textheight}{-12cm}   





\bibliographystyle{ieeetr}
\bibliography{reference}   

\begin{thebibliography}{10}

\bibitem{mountney2009dynamic}
P.~Mountney and G.-Z. Yang, ``Dynamic view expansion for minimally invasive
  surgery using simultaneous localization and mapping,'' in {\em 2009 Annual
  International Conference of the IEEE Engineering in Medicine and Biology
  Society}, pp.~1184--1187, IEEE, 2009.

\bibitem{lin2015video}
B.~Lin, Y.~Sun, X.~Qian, {\em et~al.}, ``Video-based 3d reconstruction,
  laparoscope localization and deformation recovery for abdominal minimally
  invasive surgery: a survey,'' {\em The International Journal of Medical
  Robotics and Computer Assisted Surgery}, 2015.

\bibitem{stoyanov2012stereoscopic}
D.~Stoyanov, ``Stereoscopic scene flow for robotic assisted minimally invasive
  surgery,'' in {\em International Conference on Medical Image Computing and
  Computer-Assisted Intervention}, pp.~479--486, Springer, 2012.

\bibitem{haouchine2015monocular}
N.~Haouchine, J.~Dequidt, M.-O. Berger, and S.~Cotin, ``Monocular 3d
  reconstruction and augmentation of elastic surfaces with self-occlusion
  handling,'' {\em IEEE transactions on visualization and computer graphics},
  vol.~21, no.~12, pp.~1363--1376, 2015.

\bibitem{malti2011template}
A.~Malti, A.~Bartoli, and T.~Collins, ``Template-based conformal
  shape-from-motion from registered laparoscopic images.,'' in {\em MIUA},
  vol.~1, p.~6, 2011.

\bibitem{grasa2011ekf}
O.~G. Grasa, J.~Civera, and J.~Montiel, ``Ekf monocular slam with
  relocalization for laparoscopic sequences,'' in {\em Robotics and Automation
  (ICRA), 2011 IEEE International Conference on}, pp.~4816--4821, IEEE, 2011.

\bibitem{lin2013simultaneous}
B.~Lin, A.~Johnson, X.~Qian, J.~Sanchez, and Y.~Sun, ``Simultaneous tracking,
  3d reconstruction and deforming point detection for stereoscope guided
  surgery,'' in {\em Augmented Reality Environments for Medical Imaging and
  Computer-Assisted Interventions}, pp.~35--44, Springer, 2013.

\bibitem{du2015robust}
X.~Du, N.~Clancy, {\em et~al.}, ``Robust surface tracking combining features,
  intensity and illumination compensation,'' {\em International Journal of
  Computer Assisted Radiology and Surgery}, vol.~10, no.~12, pp.~1915--1926,
  2015.

\bibitem{newcombe2011kinectfusion}
R.~A. Newcombe, S.~Izadi, {\em et~al.}, ``Kinectfusion: Real-time dense surface
  mapping and tracking,'' in {\em Mixed and Augmented Reality (ISMAR), 2011
  10th IEEE International Symposium on}, pp.~127--136, IEEE, 2011.

\bibitem{newcombe2015dynamicfusion}
R.~A. Newcombe, D.~Fox, and S.~M. Seitz, ``Dynamicfusion: Reconstruction and
  tracking of non-rigid scenes in real-time,'' in {\em Proceedings of the IEEE
  Conference on Computer Vision and Pattern Recognition}, pp.~343--352, 2015.

\bibitem{innmann2016volumedeform}
M.~Innmann, M.~Zollh{\"o}fer, M.~Nie{\ss}ner, C.~Theobalt, and M.~Stamminger,
  ``Volumedeform: Real-time volumetric non-rigid reconstruction,'' in {\em
  European Conference on Computer Vision}, pp.~362--379, Springer, 2016.

\bibitem{dou2016fusion4d}
M.~Dou, S.~Khamis, {\em et~al.}, ``Fusion4d: real-time performance capture of
  challenging scenes,'' {\em ACM Transactions on Graphics (TOG)}, vol.~35,
  no.~4, p.~114, 2016.

\bibitem{maier2013optical}
L.~Maier-Hein, P.~Mountney, A.~Bartoli, H.~Elhawary, D.~Elson, A.~Groch,
  A.~Kolb, M.~Rodrigues, J.~Sorger, S.~Speidel, {\em et~al.}, ``Optical
  techniques for 3d surface reconstruction in computer-assisted laparoscopic
  surgery,'' {\em Medical image analysis}, vol.~17, no.~8, pp.~974--996, 2013.

\bibitem{sorkine2007rigid}
O.~Sorkine and M.~Alexa, ``As-rigid-as-possible surface modeling,'' in {\em
  Symposium on Geometry Processing}, vol.~4, pp.~109--116, 2007.

\bibitem{billings2012system}
S.~Billings, N.~Deshmukh, {\em et~al.}, ``System for robot-assisted real-time
  laparoscopic ultrasound elastography,'' in {\em SPIE Medical Imaging},
  pp.~83161--83161, International Society for Optics and Photonics, 2012.

\bibitem{zhang2017autonomous}
L.~Zhang, M.~Ye, P.~Giataganas, M.~Hughes, and G.-Z. Yang, ``Autonomous
  scanning for endomicroscopic mosaicing and 3d fusion,'' in {\em Robotics and
  Automation (ICRA), 2017 IEEE International Conference on}, pp.~3587--3593,
  IEEE, 2017.

\bibitem{volumedeformurl}
M.~Innmann, M.~Zollh{\"o}fer, {\em et~al.}, ``Volumedeform: Real-time
  volumetric non-rigid reconstruction - eccv 2016.''
  \url{https://www.youtube.com/watch?v=khthUS7KVY4}, 2016.

\bibitem{uijlings2009real}
J.~R. Uijlings, A.~W. Smeulders, and R.~J. Scha, ``Real-time bag of words,
  approximately,'' in {\em Proceedings of the ACM international Conference on
  Image and Video Retrieval}, p.~6, ACM, 2009.

\bibitem{sumner2007embedded}
R.~W. Sumner, J.~Schmid, and M.~Pauly, ``Embedded deformation for shape
  manipulation,'' {\em ACM Transactions on Graphics (TOG)}, vol.~26, no.~3,
  p.~80, 2007.

\bibitem{giannarou2013probabilistic}
S.~Giannarou, M.~Visentini-Scarzanella, and G.-Z. Yang, ``Probabilistic
  tracking of affine-invariant anisotropic regions,'' {\em IEEE transactions on
  Pattern Analysis and Machine Intelligence}, vol.~35, no.~1, pp.~130--143,
  2013.

\bibitem{kenngott2015openhelp}
H.~Kenngott, J.~W{\"u}nscher, M.~Wagner, {\em et~al.}, ``Openhelp (heidelberg
  laparoscopy phantom): development of an open-source surgical evaluation and
  training tool,'' {\em Surgical Endoscopy}, vol.~29, no.~11, pp.~3338--3347,
  2015.

\bibitem{song20163d}
J.~Song, J.~Wang, L.~Zhao, S.~Huang, and G.~Dissanayake, ``3d shape recovery of
  deformable soft-tissue with computed tomography and depth scan,'' in {\em
  Australasian Conference on Robotics and Automation (ACRA)}, pp.~117--126,
  ARAA, 2016.

\end{thebibliography}

\end{document}